\documentclass[11pt,a4paper]{article}
\usepackage[hyperref]{emnlp2020}

\usepackage{times}
\usepackage{latexsym}

\usepackage{microtype}

\usepackage[font=normalsize,position=bottom]{subfig}
\usepackage{graphicx}

\usepackage{diagbox}
\usepackage{url}
\usepackage{multirow}

\usepackage{arydshln}
\usepackage{amsmath}

\usepackage{algorithm}
\usepackage{algorithmic}
\usepackage{tipa}

\usepackage[normalem]{ulem}

\newcommand\blfootnote[1]{%
  \begingroup
  \renewcommand\thefootnote{}\footnote{\noindent #1}%
  \addtocounter{footnote}{-1}%
  \endgroup
}

\title{Dynamic Data Selection and Weighting for Iterative Back-Translation}

\author{Zi-Yi Dou\textsuperscript{\textipa{Z}} \ \  Antonios Anastasopoulos\textsuperscript{\textipa{D},$\dagger$} \ \  Graham Neubig\textsuperscript{\textipa{Z}} \\
  \textsuperscript{\textipa{Z}} Language Technologies Institute, Carnegie Mellon University \\
  \textsuperscript{\textipa{D}} Department of Computer Science, George Mason University\\
  {\tt \{zdou,gneubig\}@cs.cmu.edu \ \ antonis@gmu.edu}}
\date{}
\definecolor{mypink}{rgb}{0.99, 0.4, 0.7}

\aclfinalcopy

\begin{document}
\maketitle

\begin{abstract}
Back-translation has proven to be an effective method to utilize monolingual data in neural machine translation (NMT), and iteratively conducting back-translation can further improve the model performance. Selecting which monolingual data to back-translate is crucial, as we require that the resulting synthetic data are of high quality \textit{and} reflect the target domain. To achieve these two goals, data selection and weighting strategies have been proposed, with a common practice being to select samples close to the target domain but also dissimilar to the average general-domain text. In this paper, we provide insights into this commonly used approach and generalize it to a dynamic curriculum learning strategy, which is applied to iterative back-translation models. In addition, we propose weighting strategies based on both the current quality of the sentence and its improvement over the previous iteration. We evaluate our models on domain adaptation, low-resource, and high-resource MT settings and on two language pairs. Experimental results demonstrate that our methods achieve improvements of up to~1.8 BLEU points over competitive baselines.\blfootnote{$\dagger$: Work completed while at Carnegie Mellon University.}\footnote{Code: ~\url{https://github.com/zdou0830/dynamic\_select\_weight}.}
\end{abstract}
\section{Introduction}

\begin{figure}[t]
\centering
\includegraphics[width=0.45\textwidth]{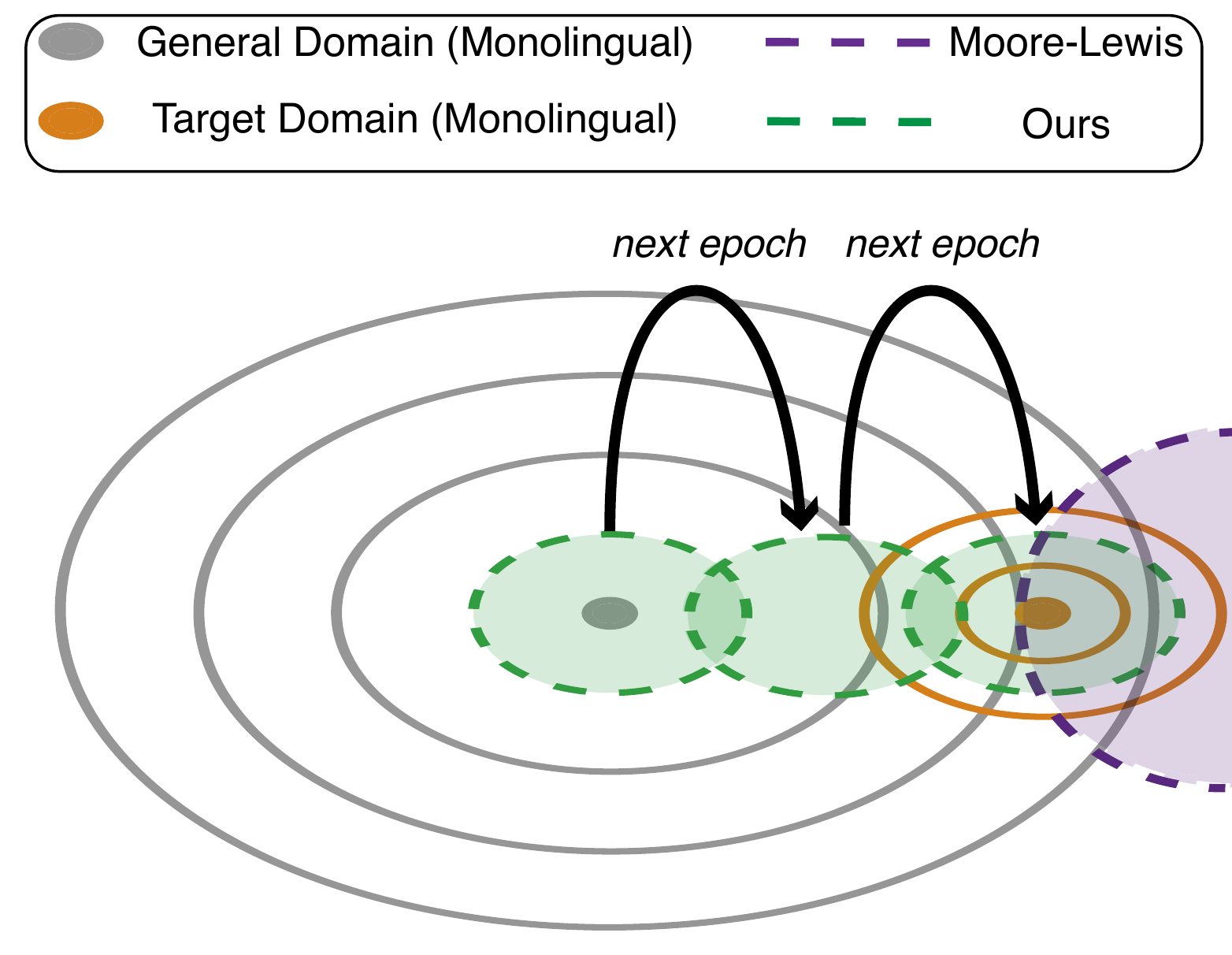}
\caption{The \newcite{moore2010intelligentselection} data selection strategy for domain adaptation constantly selects the same set of sentences which cannot well represent the target domain. Our approach, instead, selects different subsets of sentences at each epoch and we gradually shift from selecting samples from the general-domain distribution to samples from the target distribution.} 
\label{fig:fig1}
\end{figure}



Back-translation~\citep{sennrich2016neural} is an effective strategy for improving the performance of neural machine translation~(NMT) using monolingual data, delivering impressive gains over already competitive NMT models~\citep{edunov2018understanding}.
The strategy is simple: given monolingual data in the target language, one can use a translation model in the opposite of the desired translation direction to \textit{back-translate} the monolingual data, effectively synthesizing a parallel dataset, which is in turn used to train the final translation model.
Further improvements can be obtained by iteratively repeating this process~\citep{hoang2018iterative} in both directions. 

However, not all monolingual data are equally important. An envisioned downstream application is very often characterized by a unique data distribution. In such cases of domain shift, back-translating target domain data can be an effective strategy~\citep{hu2019domain} for obtaining a better in-domain translation model. 
One common strategy is to select samples that are both~(1) close to the target distribution and~(2) dissimilar to the average general-domain text~\citep{moore2010intelligentselection}. However, as depicted in Figure~\ref{fig:fig1}, this method is not ideal because the second objective could bias towards the selection of sentences far from the center of the target distribution, potentially leading to selecting a non-representative set of sentences.


Even if we could select all in-domain monolingual data, the back-translation model has not been trained on in-domain parallel data and thus the back-translated data will be of poor quality.
As we demonstrate in the experiments, the quality of the back-translated data can have a large influence on the final model performance.
To achieve the two goals of both selecting target-domain data and back-translating them with high quality, in this paper, we propose a method to combine dynamic data selection with weighting strategies for iterative back-translation. 
Specifically, the dynamic data selection selects subsets of sentences from a monolingual corpus \emph{at each training epoch}, gradually transitioning from selecting general-domain data to choosing target-domain sentences. The gradual transition ensures that the back-translation model of each iteration can adequately translate the selected sentences, as they are close to the distribution of its current training data.
We also assign weights to the back-translated data that reflect their quality, which further reduces the effect of potential noise due to low quality translations.
The proposed data selection and weighting strategies are complementary to each other, as the former focuses on domain information while the latter emphasizes the quality of sentences.


We investigate the performance of our methods in domain adaptation, low-resource and high-resource MT settings and on German-English and Lithuanian-English datasets. Our strategies demonstrate improvements of up to~1.8 BLEU points over a competitive iterative back-translation baseline and up to 1.2 BLEU points over the best static data selection strategies. In addition, our analysis reveals that the selected samples can represent the target distribution well and that the weighting strategies are effective in noisy settings.

\section{Background: Back-Translation}
\vspace{-2pt}
Back-translation~\cite{sennrich2016improving} has proven to be an effective way of utilizing monolingual data for machine translation.
Given a parallel training corpus $\mathbf{D}_{FE}$, we first train a target-to-source machine translation model $\mathbf{M}_{EF}$.
Then, we use the pre-trained model $\mathbf{M}_{EF}$ to translate a target language monolingual corpus $\mathbf{D}_{E}$ to the source language and obtain a synthetic parallel corpus $(\mathbf{D}_{F}', \mathbf{D}_E)$.
Last, we concatenate back-translated data $(\mathbf{D}_{F}', \mathbf{D}_E)$ with the original parallel corpus $\mathbf{D}_{FE}$ to train a source-to-target model $\mathbf{M}_{FE}$.

 \begin{algorithm}[t]
\caption{\label{alg:main} Iterative Back-Translation}

\begin{algorithmic}
    \REQUIRE{Monolingual corpora $\mathbf{D}_{F}$ and $\mathbf{D}_{E}$}
    \ENSURE{Translation models $\mathbf{M}_{FE}$ and $\mathbf{M}_{EF}$}
    \WHILE{$\mathbf{M}_{FE}$ and $\mathbf{M}_{EF}$ have not converged}
        \FORALL{ batches $(\mathbf{B}_{F}, \mathbf{B}_{E})$  in $(\mathbf{D}_{F}, \mathbf{D}_{E})$}
            \STATE{Translate $\mathbf{B}_F$ into $\mathbf{B}_E'$ using $\mathbf{M}_{FE}$}
            \STATE{Translate $\mathbf{B}_E$ into $\mathbf{B}_F'$ using $\mathbf{M}_{EF}$}
            \STATE{Train $\mathbf{M}_{FE}$ with $(\mathbf{B}_F', \mathbf{B}_E)$}
            \STATE{Train $\mathbf{M}_{EF}$ with $(\mathbf{B}_E', \mathbf{B}_F)$}
        \ENDFOR
	\ENDWHILE
\end{algorithmic}
\end{algorithm}


The success of back-translation has motivated researchers to investigate and extend the method~\citep{he2016dual,Zheng2020MirrorGenerativeNM}. 
~\newcite{hoang2018iterative} propose to use iterative back-translation and achieve improvements over previous state-of-the-art models. As shown in Algorithm~\ref{alg:main}, at each training step, a batch of monolingual sentences is sampled from one language and back-translated to the other language. The back-translated data is utilized to train the model in the other direction. The process is repeated in both directions. 
 
 \vspace{-2pt}

\section{Methods}
In our setting, we are given two MT models $\mathbf{M}_{FE}$ and $\mathbf{M}_{EF}$ pretrained on parallel data $\mathbf{D}_{FE}$, and both source and target monolingual corpora $\mathbf{D}_{F}$ and $\mathbf{D}_{E}$.
The goal is to select and weight samples from the two monolingual corpora for back-translation, in order to best improve the performance of the two translation models.
\begin{figure*}[t]
\centering
\includegraphics[width=1.0\textwidth]{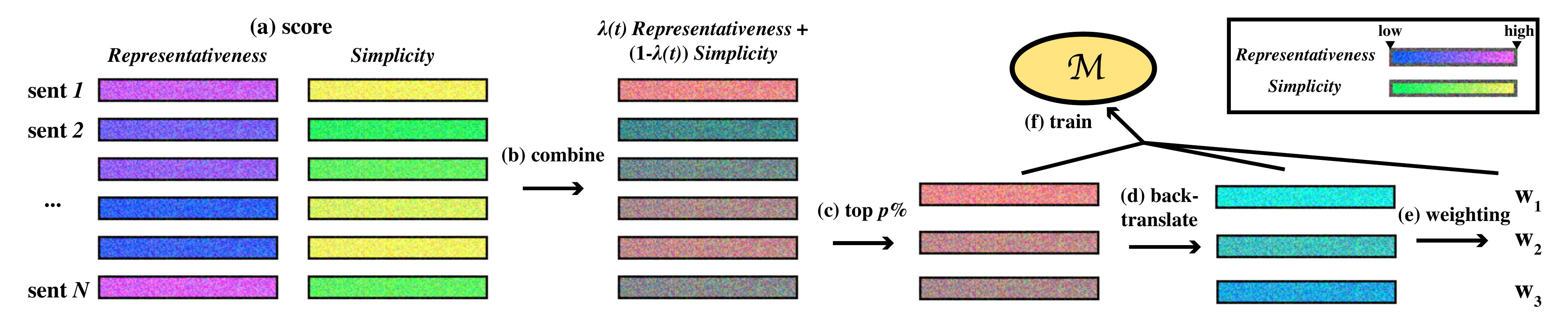}
\caption{Main procedure of our algorithm. We first compute the representative and simplicity scores for all the monolingual sentences (a). At each training epoch $t$, we combine the two scores (b) and select the top $p\%$ monolingual sentences (c). After back-translating the selected sentences from the source side to the target side (d), we then perform data weighting on the back-translated samples (e) and train the model with the weighted back-translated sentences (f).} 
\label{fig:main}
\end{figure*}
\subsection{Data Selection Strategies}
We first describe a commonly used \textit{static} selection strategy, and then illustrate our \textit{dynamic} approach.
\subsubsection{The~\newcite{moore2010intelligentselection} Method}
A common approach for data selection is the \citet{moore2010intelligentselection} method (and extensions, e.g.~\citet{axelrod2011adaptation,duh2013adaptation,santamaria2019data}), which computes the language model cross-entropy difference for each sentence $\mathbf{s}$ in a monolingual corpus:
\begin{equation}
\label{moore}
   score(\mathbf{s}) = H_{LM_{in}}(\mathbf{s}) - H_{LM_{gen}} (\mathbf{s}),
\end{equation}
where $H_{LM_{in}}(\mathbf{s})$ and $H_{LM_{gen}} (\mathbf{s})$ represent the cross-entropy scores of $s$ measured with an in-domain and a general-domain language model (LM) respectively. Sentences with the highest scores will be selected for training.
Typically, the in-domain language model $LM_{in}$ is trained with a small set of sentences in the target domain and $LM_{gen}$ is trained with all data available. 
\subsubsection{Our Two Scoring Criteria}

Instead of static data selection, we propose a new curriculum strategy for iterative back-translation. Specifically, we measure both {\bf representativeness}, {\it i.e.} how well the sentence represents the target distribution, and {\bf simplicity}, {\it i.e.} how well the MT models can translate the sentence, of each sentence $\mathbf{s}$ in the monolingual corpus. First, we select the most simple samples for back-translation to ensure the quality of the back-translated data. As the training progresses, the model will become better at translating in-domain sentences, and we will then shift to choosing more representative examples.

Formally, at each epoch $t$, we rank the corpus according to
\begin{equation}
\label{eqn:main}
\text{score} (\mathbf{s}) = \lambda(t) \text{repr}(\mathbf{s}) + (1-\lambda(t)) \text{simp}(\mathbf{s}),
\end{equation}
where $\text{repr}(\mathbf{s})$ and $\text{simp}(\mathbf{s})$ denote the representativeness and simplicity of sentence $s$ respectively, which will be dicussed in the following sections. The term $\lambda(t)$ balances between the two criteria and is a function of the current epoch $t$. 

We adopt the square-root growing function for $\lambda$~\cite{platanios2019competence} and set 
\begin{equation}
\label{eqn:lambda}
\lambda(t) = \min(1, \sqrt{t\frac{1-c_0^2}{T}} + c_0^2),
\end{equation}
 where $c_0$ is the initial value and $T$ denotes the time after
which we solely select representative samples. $\lambda$ increases relatively quickly at first and then its acceleration will be gradually decreased as the training progresses, which is suitable for our task as at first the sentences are relatively simple and thus we will not need much time on those sentences.
\paragraph{Connections to ~\newcite{moore2010intelligentselection}.} Our proposed criteria generalize~\newcite{moore2010intelligentselection}. The first term of Equation~\ref{moore}, namely $H_{LM_{in}}(\mathbf{s})$, measures the \textbf{representativeness} of data because the in-domain LM assigns low entropy to sentences that appear frequently in the target domain. The second term $H_{LM_{gen}}(\mathbf{s})$, on the other hand, measures the \textbf{simplicity} of the sentences. If $H_{LM_{gen}}(\mathbf{s})$ is high, it is likely that some $n$-grams of the sentence $\mathbf{s}$ appear frequently in the parallel training data $\mathbf{D}_{FE}$, indicating that the MT models will likely translate the sentence well. In other words, the sentence $s$ can provide limited additional information if $H_{LM_{gen}}(\mathbf{s})$ is high. Therefore, one can view~\newcite{moore2010intelligentselection} as selecting the most representative and difficult sentences.

\subsubsection{Representativeness Metrics}
We propose three approaches to measure the sentence representativeness.
\paragraph{In-Domain Language Model Cross-Entropy (LM-in).} As in~\newcite{axelrod2011adaptation,duh2013adaptation}, we can use $H_{LM_{in}}$ to measure the representativeness of the instances. Concretely, we train a language model $LM_{in}$ with in-domain monolingual data and compute the score $\frac{1}{|\mathbf{s}|} \sum_{t=1}^{|\mathbf{s}|} \log P_{LM_{in}}  (s_t|s_{<t})$ for each sentence $\mathbf{s}$.
\paragraph{TF-IDF Scores (TF-IDF).} TF-IDF score is another criterion for data selection~\cite{kirchhoff2014submodularity}. For each sentence $\mathbf{s}$, one can compute
its term frequency and  inverse document frequency for each word.
We can thus obtain the TF-IDF vector and calculate the cosine similarity between the TF-IDF vectors of $\mathbf{s}$ and each sentence $\mathbf{s}_{in}$ in a small in-domain dataset, and treat the maximum value as its representativeness score.
\paragraph{BERT Representation Similarities (BERT).} BERT~\cite{devlin2019bert} has proven to be effective for sentence representation learning. Following the conclusion of~\newcite{pires2019multilingual}, we feed each sentence to the multilingual BERT model and average the hidden states for all the input tokens except [CLS] and [SEP] at the eighth layer to obtain the sentence representation. We then compute the cosine similarity between representations of sentence $\mathbf{s}$ in the monolingual corpus and each sentence $\mathbf{s}_{in}$ in a small in-domain set, and the maximum value is treated as the representativeness score.

\subsubsection{Simplicity Metrics}
In our experiments, we study two metrics for measuring the simplicity of sentences. Note that in the field of quality estimation for MT~\cite{specia2010machine,fonseca2019findings}, researchers have proposed several existing techniques to estimate the simplicity of sentences~\cite{turchi2014adaptive,specia-etal-2015-multi,shah-etal-2015-shef,kim-lee-2016-recurrent,kepler2019unbabel,zhou2019source,qi2019nju}, and here we select a few representative approaches.
\paragraph{General-Domain Language Model Cross-Entropy (LM-gen).} We train a language model $LM_{gen}$ with the one side of the parallel training data $\mathbf{D}_{FE}$. Then, for each sentence $\mathbf{s}$ we compute the score $\frac{1}{|\mathbf{s}|} \sum_{t=1}^{|\mathbf{s}|} \log P_{LM_{gen}}  (s_t|s_{<t})$.
\paragraph{Round-Trip BLEU (R-BLEU).} Given two pre-trained MT models $\mathbf{M}_{FE}$ and $\mathbf{M}_{EF}$, round-trip translation first translates a sentence $\mathbf{s}$ into another language using $\mathbf{M}_{FE}$ and then back-translates the result using $\mathbf{M}_{EF}$, obtaining the reconstructed sentence $\mathbf{s}'$. The BLEU score between $\mathbf{s}$ and $\mathbf{s}'$ is treated as our simplicity metric.  Similar
ideas have been applied to filter sentences of low quality~\cite{imankulova2017improving}.
\paragraph{}For both the representativeness and simplicity scores, it should be noted that they are separately normalized to [0, 1], using the equation $\frac{score(\mathbf{s}) - score_{min}}{score_{max}-score_{min}}$, where $score_{max}$ and $score_{min}$ are the maximum and minimum scores.
\subsection{Weighting Strategies}
Next, we illustrate how we perform data weighting on the back-translated data.
\subsubsection{Measuring the Current Quality}
As general translation models could perform poorly on the in-domain data, we need ways to measure the current quality of the back-translated sentences in order to down-weight examples of poor quality.
\paragraph{Encoder Representation Similarities (Enc).} We feed the source sentence $\mathbf{x}$ and the target sentence $\mathbf{y}$ to the encoders of $\mathbf{M}_{FE}$ and $\mathbf{M}_{EF}$ respectively, and average the hidden states at the final layer to obtain the representations $enc_{FE}(\mathbf{x})$ and $enc_{EF}(\mathbf{y})$. The cosine similarity between them is treated as the quality metric.
\paragraph{Agreement Between Forward and Backward Models (Agree).} Inspired by~\newcite{junczys2018dual}, the second approach utilizes the agreement of the two translation models. For each sentence pair $(\mathbf{x}, \mathbf{y})$, we compute the conditional probability $H_{FE}(\mathbf{y}|\mathbf{x})$ and $H_{EF}(\mathbf{x}|\mathbf{y})$, then exponentiate the absolute value between them  $\exp(-(|H_{FE}(\mathbf{y}|\mathbf{x})-H_{EF}(\mathbf{x}|\mathbf{y})|))$. Intuitively, the back-translated sentences are of poor quality if there are huge disagreements between the two models.
\subsubsection{Measuring Quality Improvements}
In domain adaptation, it is natural that at first the in-domain sentences are poorly translated. As training progresses, however, the quality should be improved. We therefore propose a metric to measure the improvement in translation quality and combine it with the current quality metric, in order to encourage the inclusion of in-domain sentences where the translation qualities have improved.

Specifically, every time we obtain the quality score of sentence $\mathbf{s}$, we store it, then the next time we come across the same sentence, we can compare the new quality score with the previous one:
$$
\text{Imp}(\mathbf{s}) = clip(\frac{\text{current\_quality} (\mathbf{s})}{\text{previous\_quality}(\mathbf{s})}, w_{low}, w_{high}),
$$ where the clipping function limits the weights to a reasonable range. We set $(w_{low}, w_{high})$ to $(\frac{1}{2}, 2)$.

 \begin{table}[t]
  \centering
   \resizebox{0.5\textwidth}{!}{
  \begin{tabular}{c|c|c|c|c}
   \multicolumn{1}{c|}{\multirow{3}{*}{\bf Method} } &\multicolumn{4}{c}{WMT}  \\
   \cline{2-5}
   & \multicolumn{2}{c|}{LAW} & \multicolumn{2}{c}{MED}\\
   \cline{2-5}
  & de-en & en-de & de-en & en-de \\
   \hline
   \hline
   \multicolumn{5}{l}{\it Baseline} \\
  \hline
   Base & 31.25 & 24.44 & 34.43 & 26.59 \\
   Back & 35.90 & 26.33 & 42.42 & 33.98\\
    Ite-Back  & 37.69 & 27.81 & 44.08 & 35.65 \\
    \citet{zhang2019curriculum} & 37.70 & 27.87 & 44.25 & 36.01 \\
   \hline
   \hline
  \multicolumn{5}{l}{\it Best Selection} \\
  \hline
  TF-IDF & 38.26* & 28.35* & 44.26 & 35.82 \\
  \hline
   \hline
  \multicolumn{5}{l}{\it Best Curriculum} \\
  \hline
  TF-IDF + R-BLEU & \bf 39.11* &  28.93* &44.91* &36.19* \\
  \hline
  \hline
  \multicolumn{5}{l}{\it Best Weighting} \\
  \hline
  Enc & 38.20* & 28.15* & 44.28* & 35.52  \\
    Enc-Imp &  38.13* & 27.97 & 44.46* & 35.77 \\
  \hline
  \hline
  \multicolumn{5}{l}{\it Best Curriculum + Best Weighting} \\
  \hline
Curri+Enc &  38.87 & \bf 29.04 & \bf 45.46* & 36.34 \\
Curri+Enc-Imp & 38.75 & 28.89 & \bf 45.46* & \bf 36.45* \\
\hline
    \end{tabular}
    }
    \caption{ \label{tab:main1} Translation accuracy (BLEU~\cite{papineni2002bleu}) in the domain adaptation setting. The first and second rows list source and target domains respectively. The third row lists the translation directions. We report the best-performing models of only using selection strategies ({\it ``Best Selection'')}, only using curriculum strategies ({\it ``Best Curriculum''}), only using weighting strategies ({\it ``Best Weighting'' }) and using both the best curriculum and weighting strategies ({\it ``Best Weighting + Best Weighting'' }). ``Enc-Imp'' indicates both the encoder representation similarities and the quality improvement metrics are used for weighting. The highest scores are in {\bf bold} and $*$ indicates statistical significance compared with the best baseline ($p <0.05$).}
  \end{table}

\subsection{Overall Algorithm: Combining Curriculum and Weighting Strategies}
Our final algorithm is shown in Figure~\ref{fig:main}. At each epoch, we compute the score for each sentence in monolingual corpora using Equation~\ref{eqn:main} and select the top $p$\% of sentences, where $p$ is a hyper-parameter. Afterwards, we perform back-translation and data weighting on the selected data, then use the back-translated data to train the translation model. The process will be repeated iteratively for both directions, with $\lambda$ increased at each training epoch.

\begin{table}[t]
  \centering
   \resizebox{0.5\textwidth}{!}{
  \begin{tabular}{c|c|c|c|c}
   \multicolumn{1}{c|}{\multirow{3}{*}{\bf Method} } &\multicolumn{4}{c}{WMT}  \\
   \cline{2-5}
   & \multicolumn{2}{c|}{LAW} & \multicolumn{2}{c}{MED}\\
   \cline{2-5}
  & de-en & en-de & de-en & en-de \\
  \hline
  \hline
   \multicolumn{5}{l}{\it Baseline} \\
  \hline
  Ite-Back  & 37.69 & 27.81 & 44.08 & 35.65 \\
  \hline
  \hline
  \multicolumn{5}{l}{\it Selection} \\
  \hline
    BERT  &  37.84 & 28.12 & 44.17 & 35.68\\
  LM-diff & 37.91 & 27.77 & \bf 44.59 & \bf 36.00\\
  LM-in & 38.23 & 28.29 & 44.25 & 34.98 \\
  TF-IDF & \bf 38.26 & \bf 28.35 & 44.26 & 35.82 \\
  \hline
   \hline
  \multicolumn{5}{l}{\it Weighting} \\
  \hline
  Enc& \bf 38.20 &\bf 28.15 & 44.28 & 35.52  \\
    Enc-Imp &  38.13 & 27.97 & \bf 44.46 & \bf 35.77 \\
    \hdashline
    Agree &  37.41 & 27.70 & 44.04 & 35.70 \\
  Agree-Imp & 37.42 & 27.78&  44.30 & 35.37 \\
  \hline
  \hline
    \multicolumn{5}{l}{\it Curriculum} \\
  \hline
  LM-in+ LM-gen &  38.26 & 28.51 &44.68 & 34.90 \\
  TF-IDF + LM-gen & 38.67 & 28.67 & 44.90 & 35.49 \\
  TF-IDF + R-BLEU & \textit{\textbf{ 39.11}} & \textit{\textbf{ 28.93}} & \textit{\textbf{44.91}} & \textit{\textbf{ 36.19}} \\
  \hline
    \end{tabular}
    }
    \caption{ \label{tab:comp1} Comparisons of different metrics in domain adaptation. The highest scores in each section are in {\bf bold} and the overall highest scores are in \textit{\textbf{bold italic}}.} 
  \end{table}

\section{Experiments on Domain Adaptation}
We first conduct experiments in the domain adaptation setting, where we adapt models from a general domain to a specific domain.
\subsection{Setup}
\label{subsec:setup}
\paragraph{Datasets.} We first train the translation models with (general-domain) WMT-14 German-English dataset, consisting of about 4.5M training sentences, then perform iterative back-translation with (in-domain) law or medical OPUS monolingual data~\cite{tiedemann2012parallel}. We de-duplicate the law and medical parallel training data, divide them into two halves and obtain 250K and 200K comparable yet non-parallel sentences respectively in both languages to obtain the monolingual corpora. The development and test sets contain 2K sentences in each domain. Byte-pair encoding~\cite{sennrich2016neural} is applied with 32K merge operations. The general-domain and in-domain language models are trained on the WMT training data and the OPUS monolingual data respectively. The OPUS development sets are used to compute the TF-IDF and BERT representativeness scores.

\paragraph{Models.} We implement our approaches upon the Transformer~\cite{vaswani2017attention}. Both the encoder and decoder consist of 6 layers and the hidden size is set to 512. For the translation models, weights of the top 4 layers of the encoders and bottom 4 layers of the decoders are shared between forward and backward models. We also tie the source and target word embeddings. We build 5-gram language models with modified Kneser-Ney
smoothing using KenLM~\cite{heafield2011kenlm}.


\paragraph{Hyper-Parameters.} $c_0$ and $T$ in Equation~\ref{eqn:lambda} are set to 0.1 and 5. We select 30\% of the sentences with the highest score at each epoch for our curriculum methods and 50\% of the sentences for the static data selection baselines.

\begin{table}[t]
  \centering
  \begin{tabular}{c|c|c|c}
 \bf Method     & de-en & en-de  & Avg. $\Delta$  \\
   \hline
  \hline
   Ite-Sampling & 34.93 & 26.72 & - \\
   Ite-Sampling + Enc& 35.67 & 27.76 & +0.89\\
   \hdashline
   Ite-Greedy & 37.69 & 27.81 & -\\
   Ite-Greedy+Enc & 38.20 & 28.15 & +0.43 \\
    \hdashline
   Ite-Beam & 37.53 & 28.25 &- \\
   Ite-Beam+Enc & 37.76 & 28.25 & +0.12 \\
   \hline
    \end{tabular}
    \caption{ \label{tab:effect} Noise in back-translated data can degrade the model performance and our weighting strategies (Enc) benefit the most in noisy settings.}
  \end{table}

\begin{table*}[t]
  \centering
   \resizebox{1.0\textwidth}{!}{%
  \begin{tabular}{c||c|c|c}
  & Back-Translated Sentence & Weight & BLEU \\
  \hline
  \hline
  Source & - wenn der Viehhalter seinen Betrieb einem Nachfolger bis zum dritten Verwandtschaftsgrad übergibt ; & - &- \\
  Reference & - when the farmer gives over his farm to his family successor up to the third degree of relationship , & - & -\\
  \hline
  Ite-5K & - if the livestock farmer hands over his holding to a successor up to the third degree of kinship ;   & 0.550 & 0.353 \\
  Ite-10K & - when the livestock farmer passes his holding to a successor up to the third degree of kinship ;  & 0.572 & 0.383 \\
  Ite-15K & - when the livestock farmer gives his holding to a successor up to the third degree of kinship ; & 0.585 & 0.402 \\
  \hline
  \hline
  Source & folgerichtig sollte dies auch auf Antisubventionsuntersuchungen zutreffen . & -& - \\
  Reference & the same principles should logically apply to anti - subsidy investigations .  & -& -\\
  \hline
  Ite-5K & this should also be followed up by anti - subsidy investigations . & 0.389 & 0.331 \\
  Ite-10K  & it should also be folly to apply to anti - subsidy investigations . &0.403 & 0.486   \\
  Ite-15K &it should also be folly true to apply to anti - subsidy investigations . &  0.397 & 0.447\\
  \hline
  \end{tabular}
  }
    \caption{\label{tab:weight_exam} Examples of our weighting strategy (Enc). We use our model (Curri+Enc) at the 5K-, 10K-, 15K-th iterative back-translation step to weight sentences. The assigned weights correlate well with the BLEU scores.}
  \end{table*}

\subsection{Results}
We compare our dynamic curriculum and weighting methods with three baselines: the iterative back-translation baseline, a baseline trained with only data selection strategies, a baseline trained with only data weighting strategies. The results with the best-performing representativeness and simplicity metrics (TF-IDF and R-BLEU, respectively) in the domain adaptation setting are listed in Table~\ref{tab:main1}.

\paragraph{Iterative Back-Translation.} The iterative back-translation method is rather competitive, as it improves over the unadapted baseline by~9.6 BLEU and simple back-translation by 1.8 BLEU points. 

\paragraph{Selection Strategies.} We can see from the table that the best-performing selection strategies, namely selecting sentences with high TF-IDF scores, is generally effective and can improve the baseline by about 0.5 BLEU points.

\paragraph{Curriculum and Weighting Strategies.} Both our curriculum and weighting strategies outperform the unadapted and the iterative back-translation models, as well as the curriculum method proposed in~\citet{zhang2019curriculum}, with our curriculum learning method achieving better performance and improving the strong iterative back-translation baseline by 1.1 BLEU points. Combining curriculum and weighting methods can further improve the performance by up to 0.5 BLEU points, demonstrating the two strategies are complementary to each other.

\begin{table}[t]
  \centering
  \begin{tabular}{@{}c||@{}c|c@{}}
  \multirow{2}{*}{\footnotesize \bf R-BLEU} & \multicolumn{2}{c}{\footnotesize \bf TF-IDF}\\
  \cline{2-3}
  &\bf \small High ($\approx1$) & \bf \small Low ($\approx0$)\\
   \hline
  \hline
 \bf \small High ($\approx1$) & \small Article 20 & \small ( 2005 / 686 / EC )  \\
  \hline
  \multirow{3}{*}{\bf \small Low ($\approx0$)}  & \multirow{1}{*}[-0.5em]{ \small{any Contracting Party }}    &
  \multirow{2}{*}[-0.2em]{\small MS Danuta}  \\
  & \small may request that & \\
  &  \multirow{1}{*}[0.5em]{\small a meeting be held .}& \multirow{1}{*}[0.5em]{ \small HÜBNER} \\
  \hline
    \end{tabular}
    \caption{\label{tab:example} Example full sentences with different TF-IDF and R-BLEU scores. R-BLEU correlates with the lengths while TF-IDF measures the domain distance.}
  \end{table}

\begin{table}[t]
  \centering
   \resizebox{\columnwidth}{!}{%
  \begin{tabular}{@{}c||c|c|c|c@{}}
   & \bf train& \bf dev &\bf  test &\bf  mono \\
   \hline
  \hline
  \multirow{3}{*}{\bf low} & \multirow{3}{*}{WMT en-de (100K)} & test2013 (3K) & test2014 (3K) & CC (1M) \\
 & & LAW (2K) &  LAW (2K) & LAW (25K)\\
 & & MED (2K) &  MED (2K) & MED (20K)\\
  \hline
  \multirow{3}{*}{\bf high} & WMT en-de (4.5M) & test2013 (3K) & test2014 (3K) & CC (10M)\\
  \cdashline{2-5}
 & \multirow{2}{*}{WMT en-lt (2M)} & \multirow{2}{*}{dev2019 (2K)} &  \multirow{2}{*}{test2019 (1K)} & News lt (5M) + \\
 &&&&  CC en (5M)\\
 \hline
    \end{tabular}
    }
    \caption{\label{tab:stats} Sources and numbers of sentences of the datasets in both low- and high- resource settings. ``CC'' refers to the CommonCrawl corpus.}
  \end{table}

  \begin{table*}[t]
  \centering
   \resizebox{1.0\textwidth}{!}{%
  \begin{tabular}{c|c|c|c|c|c|c||c|c|c|c}
   \multicolumn{1}{c|}{\multirow{3}{*}{\bf Method} } &\multicolumn{6}{c||}{WMT-low} &\multicolumn{4}{c}{WMT-high} \\
   \cline{2-11} 
   & \multicolumn{2}{c|} {News}& \multicolumn{2}{c|}{LAW} & \multicolumn{2}{c||}{MED} & \multicolumn{2}{c|} {News} & \multicolumn{2}{c} {News}\\
   \cline{2-11}
  & de-en & en-de & de-en & en-de & de-en & en-de & de-en & en-de & lt-en & en-lt\\
   \hline
  \hline
  \multicolumn{5}{l}{\it Baseline} \\
  \hline
   Base & 8.60 & 6.37 & 5.51 & 4.76 & 6.03 & 5.19 & 32.43 & 27.34 & 16.24 & 11.20\\
    Ite-Back  & 15.80 & 12.18 & 20.27 & 12.41 & 29.64 & 21.90 & 33.02 & 27.82 & 19.44 & 12.41\\
    \hline
  \hline
  \multicolumn{5}{l}{\it Best Selection} \\
  \hline
  Select& 15.44 & 12.09 & 21.19* & 12.70 & 30.84* & 21.97 & 32.89 & 27.97 & 19.52 & 12.20\\
  \hline
  \hline
  \multicolumn{5}{l}{\it Best Curriculum} \\
  \hline
  Curri & \bf 16.45* & 12.61* &  21.53* & 12.97* & \bf 31.22* & 21.71 & 33.34* & \bf 28.12* & 19.82* & 12.48\\
  \hline
  \hline
  \multicolumn{5}{l}{\it Weighting} \\
  \hline
  Enc & 16.03 & 12.59* & 20.24 & 12.55 & 29.95* & \bf 22.18* & 32.80 & 28.03 & 19.64 & 12.46\\
  Agree &   15.80 &12.55* & 20.76*  & 12.85* & 29.96* & 21.69 & 32.80 & 28.00 & 19.53 & \bf 12.66 \\
  \hline
  \hline
  \multicolumn{5}{l}{\it Best Curriculum + Weighting} \\
  \hline
  Curri+Enc &  16.24& \bf 12.70&  21.30 & 12.99 &  30.82 & 21.56 & 33.21 & 27.97 & \bf 20.05 & 12.50\\
Curri+Enc-Diff & 16.13 &12.65 &   21.80*  & 13.18 & 30.73 & 21.58  & 33.15 & 28.02 & 19.51 & 12.39 \\
Curri+Agree & 16.23 & 12.40 &  21.83* & 13.13 & 30.78 & 21.66 & 33.10 & 27.99 & 19.73 & 12.48 \\
Curri+Agree-Diff & 16.20 & 12.61& \bf 22.06* & \bf 13.28* & 30.75 & 21.30 & \bf 33.45 & 27.91 & 19.48 & 12.42 \\
\hline
    \end{tabular}}
    \caption{ \label{tab:main2} Translation accuracy (BLEU) in low-resource and high-resource scenarios. The first and second row list the source and target domains. The third row lists the translation directions. The highest scores are in {\bf bold} and $*$ indicates statistical significance compared with the best baseline  ($p <
0.05$). }
  \end{table*}

\subsection{Choices of Metrics}

We examine different choices of representativeness and simplicity metrics. The performance of different models is listed in Table~\ref{tab:comp1}. 

\paragraph{Representativeness Metrics.} All data selection strategies outperform the baseline, with TF-IDF, LM-diff, and BERT metrics exhibiting fairly robust performance in all settings. Due to its simplicity, we choose TF-IDF for experiments where a good in-domain development set is available.

\paragraph{Data Weighting Strategies.} The agreement-based weighting method (``Agree'') performs slightly worse than the encoder-similarity weighting strategy (``Enc''), probably because the two languages are similar and thus encoders with shared parameters can accurately measure the data quality.

\paragraph{Curriculum Strategies.} Table~\ref{tab:comp1} demonstrates that TF-IDF is a better metric than other representativeness metrics in both static and dynamic data selection settings. Also, the round-trip BLEU score can be better at measuring the simplicity of sentences than LM-gen. Last, by comparing the Moore-Lewis method (``LM-diff'') with our curriculum strategy (``LM-in+LM-gen''), we can see that our method outperform Moore-Lewis method in 3 out of 4 settings. 

\subsection{Analysis}
Next, we investigate how noise in the back-translated data impacts the model performance, how many sentences we should select, and if our weighting methods assign weights appropriately.

\paragraph{Effect of Back-Translation Quality.} We try to generate the back-translated data using sampling, greedy search and beam search for iterative back-translation and the results are listed in Table~\ref{tab:effect}. We find that the sampling method significantly degrades the model performance, as it introduces more noise than other approaches, demonstrating that noise can have a negative impact in domain adaptation settings. The conclusion is similar to the findings in low-resource settings~\newcite{edunov2018understanding}. In addition, we find that our weighting strategies are more beneficial in noisy settings.

\begin{figure}[t]
\centering
\includegraphics[width=0.35\textwidth]{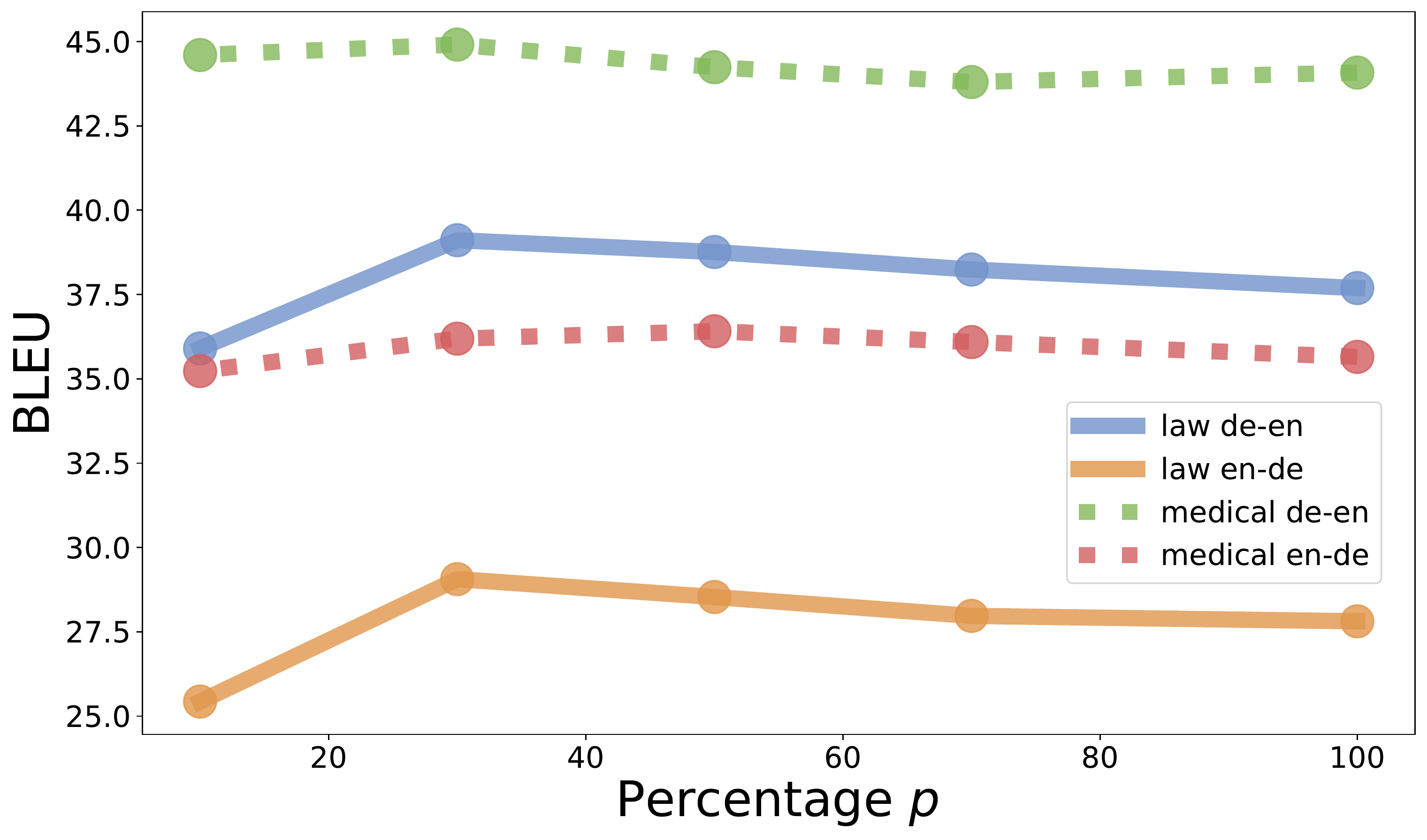}
\caption{While the model is relatively robust to the number of selected sentences at each epoch, selecting too many or too few sentences can be harmful.}
\label{fig:ablation_p}
\end{figure}
\paragraph{Effect of the Percentage $p$.} We test how many sentences should be selected at each epoch for our curriculum strategies. As shown in Figure~\ref{fig:ablation_p}, selecting 30\% of the monolingual sentences achieves the best performance in general. Selecting fewer samples can discard valuable information whereas choosing more instances can introduce more noise.

\paragraph{Weighting Examples.} We use our model (Curri+Enc) to back-translate some sentences from the monolingual corpus and Table~\ref{tab:weight_exam} shows the weights our models assign at different training stages. In this example, the assigned weights correlate well with the BLEU scores, demonstrating our methods can perform weighting appropriately in some cases.

\subsection{Characteristics of the Selected Data}

\begin{figure}[t]
\centering
\includegraphics[width=0.45\textwidth]{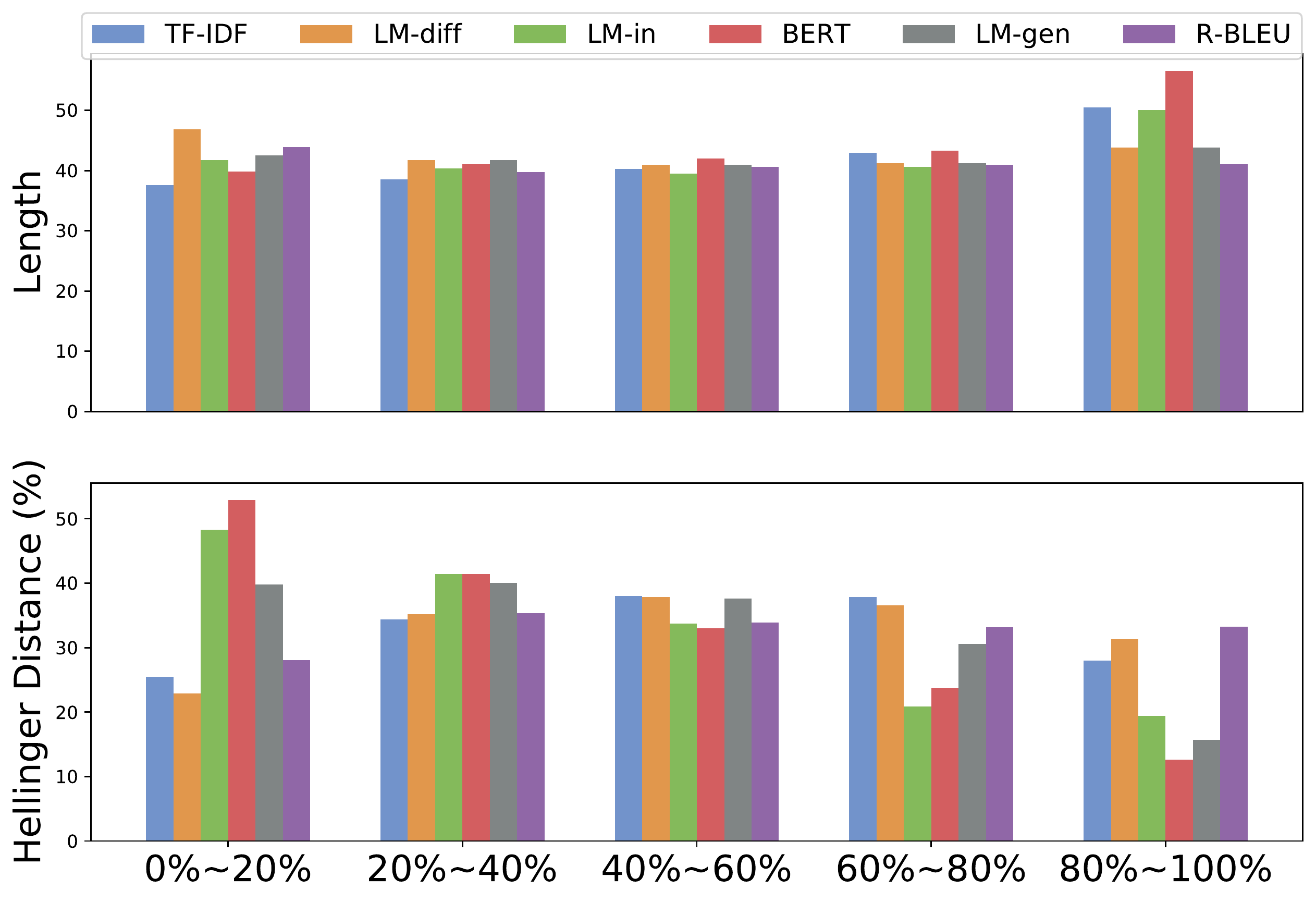}
\caption{Length and Hellinger distance of sentences in each bucket selected with different metrics.}
\label{fig:charc}
\end{figure}

In this part, we investigate certain characteristics of the selected samples.
\paragraph{Lengths.} Figure ~\ref{fig:charc} shows the average lengths of the selected sentences in each bucket. We can see that 1) both LM-in and BERT favor long sentences, with one possible explanation being that those sentences are more likely to contain in-domain words; 2) TF-IDF does not share this feature, likely due to the IDF term;  3) sentences with high R-BLEU scores are generally short, likely because NMT models are bad at translating long sentences.

\paragraph{Unigram Distribution Distance.} We also compute the unigram distribution distance using the Hellinger distance. Concretely, we compute the unigram distribution $P$ and $Q$ for both the selected data and the test set, and calculate
$$
\frac{1}{\sqrt{2}} \sqrt{\sum_{i=1}^V (\sqrt{p_i} - \sqrt{q_i})^2 },
$$
where $V$ is the size of the vocabulary. The larger the Hellinger distance is, the more dissimilar the two distributions are. Figure~\ref{fig:charc} shows that both TF-IDF and BERT match the test distribution well. Also, LM-in performs better than LM-diff, which confirms our hypothesis that the data selected by the Moore-Lewis method cannot adequately represent the target distribution.

\paragraph{Diversity Among Selected Data at Each Epoch.}
As our curriculum strategies dynamically select different subsets of data, here we examine how many new sentences are actually introduced at each epoch. We find that starting from the second epoch, 12.5\%, 10.4\%, 12.5\%, 18.3\%, 21.5\% of the selected sentences will be replaced at each epoch, and 52.5\% of the monolingual sentences will be selected at least once in total.

\paragraph{Examples.} Table~\ref{tab:example} shows examples of the selected sentences. Sentences with both high TF-IDF and R-BLEU scores are typically short and match the target distribution well. Sentences with high TF-IDF but low R-BLEU scores can be long and contain some out-of-vocabulary words, while sentences with low TF-IDF but high R-BLEU scores are generally short and frequently include digits and single characters. Most of the sentences with both low TF-IDF and R-BLEU scores are extremely noisy and can be safely discarded.

\section{Experiments on Low-Resource and High-Resource Scenarios}

Next, we conduct experiments in both low- and high-resource scenarios over two language pairs: Lithuanian-English and German-English.

\subsection{Setup}
Data statistics are shown in Table~\ref{tab:stats}. When the target distribution is the news domain, we train the in-domain LMs with 500K sentences from the news monolingual data. The other settings (including hyperparameters) are the same as before.



\subsection{Results}
The results are reported in Table~\ref{tab:main2}.
We find that LM-in and LM-gen is the best metric combination for curriculum strategies when the target distribution is the news domain. TF-IDF and R-BLEU as the representativeness and simplicity metrics are the best in all other settings.

\paragraph{Low-Resource Settings.} In low-resource settings, iterative back-translation can improve the baseline model by a large margin, and our curriculum strategies can still outperform the strong baseline by 1.3 BLEU points. Weighting methods also generally help and in the best case scenario, our method can improve iterative back-translation by 1.8 BLEU points. 

\paragraph{High-Resource Settings.} In high-resource settings, our curriculum strategies improve the iterative back-translation baseline by up to 0.3 BLEU points. Data weighting strategies do not always help, probably because in high-resource settings the back-translated data is already of high quality. In the best case scenario, our method outperforms iterative back-translation by 0.6 BLEU points.

\section{Related Work}
Back-translation~\cite{sennrich2016improving} has proven to be effective and several extensions of it have been proposed~\cite{he2016dual,cheng2016semi,zhang2016exploiting,xia-etal-2019-generalized}, among which iterative back-translation methods~\cite{cotterell2018explaining,hoang2018iterative,niu2018bi,Zheng2020MirrorGenerativeNM} have demonstrated strong empirical performance.
 
For domain adaptation, \newcite{moore2010intelligentselection} and \newcite{kirchhoff2014submodularity} use language model cross entropy differences and TF-IDF to select data that are similar to in-domain text respectively. ~\newcite{van2017dynamic} propose dynamic data selection strategies for machine translation models, and~\newcite{zhang2019curriculum} extend the idea to curriculum strategies. As for filtering noisy sentences, ~\newcite{junczys2018dual} propose to utilize the agreement between forward and backward translation models and
 ~\newcite{wang2019improving} propose uncertainty-based confidence estimation to improve back-translation.
 ~\newcite{wang-etal-2019-dynamically} compose dynamic domain-data selection with dynamic clean-data selection. Our methods generalize previous data selection strategies and our primary focus is to improve iterative back-translation, but our work could be extended to also include training-time dynamic data selection approaches such as the technique of \citet{wang-etal-2020-balancing}.
\section{Conclusion}
In this paper, we provide a novel insight into a widely-used data selection method~\cite{moore2010intelligentselection} and generalize it to a curriculum strategy for iterative back-translation. We also propose data weighting methods to down-weight examples of poor quality. Extensive experiments are performed to evaluate the performance of our methods; analyses reveal the selected samples can represent the target domain well and our weighting strategies benefit noisy settings the most. 
\section*{Acknowledgements}
The authors are grateful to the anonymous reviewers for their constructive comments, and to Xinyi Wang, Chunting Zhou for their valuable feedback. We also thank Amazon for providing GPU credits.

\bibliographystyle{acl_natbib}
\bibliography{emnlp2020}

\begin{thebibliography}{41}
\expandafter\ifx\csname natexlab\endcsname\relax\def\natexlab#1{#1}\fi

\bibitem[{Axelrod et~al.(2011)Axelrod, He, and Gao}]{axelrod2011adaptation}
Amittai Axelrod, Xiaodong He, and Jianfeng Gao. 2011.
\newblock \href {https://www.aclweb.org/anthology/D11-1033.pdf} {Domain
  adaptation via pseudo in-domain data selection}.
\newblock In \emph{Conference on Empirical Methods in Natural Language
  Processing (EMNLP)}.

\bibitem[{Cheng et~al.(2016)Cheng, Xu, He, He, Wu, Sun, and
  Liu}]{cheng2016semi}
Yong Cheng, Wei Xu, Zhongjun He, Wei He, Hua Wu, Maosong Sun, and Yang Liu.
  2016.
\newblock \href {https://www.aclweb.org/anthology/P16-1185.pdf}
  {Semi-supervised learning for neural machine translation}.
\newblock In \emph{Annual Meeting of the Association for Computational
  Linguistics (ACL)}.

\bibitem[{Cotterell and Kreutzer(2018)}]{cotterell2018explaining}
Ryan Cotterell and Julia Kreutzer. 2018.
\newblock \href {https://arxiv.org/pdf/1806.04402} {Explaining and generalizing
  back-translation through wake-sleep}.
\newblock \emph{arXiv}.

\bibitem[{Devlin et~al.(2019)Devlin, Chang, Lee, and
  Toutanova}]{devlin2019bert}
Jacob Devlin, Ming-Wei Chang, Kenton Lee, and Kristina Toutanova. 2019.
\newblock \href {https://www.aclweb.org/anthology/N19-1423.pdf} {Bert:
  Pre-training of deep bidirectional transformers for language understanding}.
\newblock In \emph{Meeting of the North American Chapter of the Association for
  Computational Linguistics (NAACL)}.

\bibitem[{Duh et~al.(2013)Duh, Neubig, Sudoh, and Tsukada}]{duh2013adaptation}
Kevin Duh, Graham Neubig, Katsuhito Sudoh, and Hajime Tsukada. 2013.
\newblock \href {https://www.aclweb.org/anthology/P13-2119.pdf} {Adaptation
  data selection using neural language models: Experiments in machine
  translation}.
\newblock In \emph{Annual Meeting of the Association for Computational
  Linguistics (ACL)}.

\bibitem[{Edunov et~al.(2018)Edunov, Ott, Auli, and
  Grangier}]{edunov2018understanding}
Sergey Edunov, Myle Ott, Michael Auli, and David Grangier. 2018.
\newblock \href {https://www.aclweb.org/anthology/D18-1045.pdf} {Understanding
  back-translation at scale}.
\newblock In \emph{Conference on Empirical Methods in Natural Language
  Processing (EMNLP)}.

\bibitem[{Fonseca et~al.(2019)Fonseca, Yankovskaya, Martins, Fishel, and
  Federmann}]{fonseca2019findings}
Erick Fonseca, Lisa Yankovskaya, Andr{\'e}~FT Martins, Mark Fishel, and
  Christian Federmann. 2019.
\newblock \href {https://www.aclweb.org/anthology/W19-5401.pdf} {Findings of
  the wmt 2019 shared tasks on quality estimation}.
\newblock In \emph{Conference on Machine Translation (WMT)}.

\bibitem[{He et~al.(2016)He, Xia, Qin, Wang, Yu, Liu, and Ma}]{he2016dual}
Di~He, Yingce Xia, Tao Qin, Liwei Wang, Nenghai Yu, Tie-Yan Liu, and Wei-Ying
  Ma. 2016.
\newblock \href
  {http://papers.nips.cc/paper/6469-dual-learning-for-machine-translation.pdf}
  {Dual learning for machine translation}.
\newblock In \emph{Conference on Neural Information Processing Systems
  (NeurIPS)}.

\bibitem[{Heafield(2011)}]{heafield2011kenlm}
Kenneth Heafield. 2011.
\newblock \href {https://www.aclweb.org/anthology/W11-2123.pdf} {Kenlm: Faster
  and smaller language model queries}.
\newblock In \emph{Conference on Machine Translation (WMT)}.

\bibitem[{Hoang et~al.(2018)Hoang, Koehn, Haffari, and
  Cohn}]{hoang2018iterative}
Vu~Cong~Duy Hoang, Philipp Koehn, Gholamreza Haffari, and Trevor Cohn. 2018.
\newblock \href {https://www.aclweb.org/anthology/W18-2703.pdf} {Iterative
  back-translation for neural machine translation}.
\newblock In \emph{Workshop on Neural Generation and Translation (WNGT)}.

\bibitem[{Hou et~al.(2019)Hou, Huang, Ning, Dai, and Chen}]{qi2019nju}
Qi~Hou, Shujian Huang, Tianhao Ning, Xinyu Dai, and Jiajun Chen. 2019.
\newblock \href {https://www.aclweb.org/anthology/W19-5409.pdf} {Nju
  submissions for the wmt19 quality estimation shared task}.
\newblock In \emph{Conference on Machine Translation (WMT)}.

\bibitem[{Hu et~al.(2019)Hu, Xia, Neubig, and Carbonell}]{hu2019domain}
Junjie Hu, Mengzhou Xia, Graham Neubig, and Jaime Carbonell. 2019.
\newblock \href {https://www.aclweb.org/anthology/P19-1286.pdf} {Domain
  adaptation of neural machine translation by lexicon induction}.
\newblock In \emph{Annual Meeting of the Association for Computational
  Linguistics (ACL)}.

\bibitem[{Imankulova et~al.(2017)Imankulova, Sato, and
  Komachi}]{imankulova2017improving}
Aizhan Imankulova, Takayuki Sato, and Mamoru Komachi. 2017.
\newblock \href {https://www.aclweb.org/anthology/W17-5704.pdf} {Improving
  low-resource neural machine translation with filtered pseudo-parallel
  corpus}.
\newblock In \emph{Workshop on Asian Translation (WAT)}.

\bibitem[{Junczys-Dowmunt(2018)}]{junczys2018dual}
Marcin Junczys-Dowmunt. 2018.
\newblock \href {https://www.aclweb.org/anthology/W18-64.pdf#page=918} {Dual
  conditional cross-entropy filtering of noisy parallel corpora}.
\newblock In \emph{Conference on Machine Translation (WMT)}.

\bibitem[{Kepler et~al.(2019)Kepler, Tr{\'e}nous, Treviso, Vera, G{\'o}is,
  Farajian, Lopes, and Martins}]{kepler2019unbabel}
Fabio Kepler, Jonay Tr{\'e}nous, Marcos Treviso, Miguel Vera, Ant{\'o}nio
  G{\'o}is, M~Amin Farajian, Ant{\'o}nio~V Lopes, and Andr{\'e}~FT Martins.
  2019.
\newblock \href {https://www.aclweb.org/anthology/W19-5406.pdf} {Unbabel’s
  participation in the wmt19 translation quality estimation shared task}.
\newblock In \emph{Conference on Machine Translation (WMT)}.

\bibitem[{Kim and Lee(2016)}]{kim-lee-2016-recurrent}
Hyun Kim and Jong-Hyeok Lee. 2016.
\newblock \href {https://www.aclweb.org/anthology/N16-1059} {A recurrent neural
  networks approach for estimating the quality of machine translation output}.
\newblock In \emph{Conference of the North {A}merican Chapter of the
  Association for Computational Linguistics (NAACL)}.

\bibitem[{Kirchhoff and Bilmes(2014)}]{kirchhoff2014submodularity}
Katrin Kirchhoff and Jeff Bilmes. 2014.
\newblock \href {https://www.aclweb.org/anthology/D14-1014.pdf} {Submodularity
  for data selection in machine translation}.
\newblock In \emph{Conference on Empirical Methods in Natural Language
  Processing (EMNLP)}.

\bibitem[{Moore and Lewis(2010)}]{moore2010intelligentselection}
Robert~C. Moore and William Lewis. 2010.
\newblock \href {https://www.aclweb.org/anthology/P10-2041.pdf} {Intelligent
  selection of language model training data}.
\newblock In \emph{Annual Meeting of the Association for Computational
  Linguistics (ACL)}.

\bibitem[{Neubig et~al.(2019)Neubig, Dou, Hu, Michel, Pruthi, and
  Wang}]{neubig2019compare}
Graham Neubig, Zi-Yi Dou, Junjie Hu, Paul Michel, Danish Pruthi, and Xinyi
  Wang. 2019.
\newblock \href {https://arxiv.org/pdf/1903.07926} {compare-mt: A tool for
  holistic comparison of language generation systems}.
\newblock In \emph{Meetings of the North American Chapter of the Association
  for Computational Linguistics (NAACL) Demo Track}.

\bibitem[{Niu et~al.(2018)Niu, Denkowski, and Carpuat}]{niu2018bi}
Xing Niu, Michael Denkowski, and Marine Carpuat. 2018.
\newblock \href {https://www.aclweb.org/anthology/W18-2710.pdf} {Bi-directional
  neural machine translation with synthetic parallel data}.
\newblock \emph{Workshop on Neural Generation and Translation (WNGT)}.

\bibitem[{Papineni et~al.(2002)Papineni, Roukos, Ward, and
  Zhu}]{papineni2002bleu}
Kishore Papineni, Salim Roukos, Todd Ward, and Wei-Jing Zhu. 2002.
\newblock \href {https://www.aclweb.org/anthology/P02-1040.pdf} {{BLEU}: a
  method for automatic evaluation of machine translation}.
\newblock In \emph{Annual Meeting of the Association for Computational
  Linguistics (ACL)}.

\bibitem[{Pires et~al.(2019)Pires, Schlinger, and
  Garrette}]{pires2019multilingual}
Telmo Pires, Eva Schlinger, and Dan Garrette. 2019.
\newblock \href {https://arxiv.org/pdf/1906.01502} {How multilingual is
  multilingual bert?}
\newblock \emph{arXiv}.

\bibitem[{Platanios et~al.(2019)Platanios, Stretcu, Neubig, Poczos, and
  Mitchell}]{platanios2019competence}
Emmanouil~Antonios Platanios, Otilia Stretcu, Graham Neubig, Barnabas Poczos,
  and Tom Mitchell. 2019.
\newblock \href {https://www.aclweb.org/anthology/N19-1119.pdf}
  {Competence-based curriculum learning for neural machine translation}.
\newblock In \emph{Meeting of the North American Chapter of the Association for
  Computational Linguistics (NAACL)}.

\bibitem[{Santamar{\'\i}a and Axelrod(2019)}]{santamaria2019data}
Luc{\'\i}a Santamar{\'\i}a and Amittai Axelrod. 2019.
\newblock \href {http://workshop2017.iwslt.org/downloads/O2-2-Paper.pdf} {Data
  selection with cluster-based language difference models and cynical
  selection}.
\newblock In \emph{International Conference on Spoken Language Translation
  (IWSLT)}.

\bibitem[{Sennrich et~al.(2016{\natexlab{a}})Sennrich, Haddow, and
  Birch}]{sennrich2016improving}
Rico Sennrich, Barry Haddow, and Alexandra Birch. 2016{\natexlab{a}}.
\newblock \href {https://www.aclweb.org/anthology/P16-1009.pdf} {Improving
  neural machine translation models with monolingual data}.
\newblock In \emph{Annual Meeting of the Association for Computational
  Linguistics (ACL)}.

\bibitem[{Sennrich et~al.(2016{\natexlab{b}})Sennrich, Haddow, and
  Birch}]{sennrich2016neural}
Rico Sennrich, Barry Haddow, and Alexandra Birch. 2016{\natexlab{b}}.
\newblock \href {https://www.aclweb.org/anthology/P16-1162.pdf} {Neural machine
  translation of rare words with subword units}.
\newblock In \emph{Annual Meeting of the Association for Computational
  Linguistics (ACL)}.

\bibitem[{Shah et~al.(2015)Shah, Logacheva, Paetzold, Blain, Beck, Bougares,
  and Specia}]{shah-etal-2015-shef}
Kashif Shah, Varvara Logacheva, Gustavo Paetzold, Frederic Blain, Daniel Beck,
  Fethi Bougares, and Lucia Specia. 2015.
\newblock \href {https://www.aclweb.org/anthology/W15-3041} {{SHEF}-{NN}:
  Translation quality estimation with neural networks}.
\newblock In \emph{Workshop on Statistical Machine Translation (WMT)}.

\bibitem[{Specia et~al.(2015)Specia, Paetzold, and
  Scarton}]{specia-etal-2015-multi}
Lucia Specia, Gustavo Paetzold, and Carolina Scarton. 2015.
\newblock \href {https://www.aclweb.org/anthology/P15-4020} {Multi-level
  translation quality prediction with {Q}u{E}st++}.
\newblock In \emph{Annual Meeting of the Association for Computational
  Linguistics (ACL) Demo Track}.

\bibitem[{Specia et~al.(2010)Specia, Raj, and Turchi}]{specia2010machine}
Lucia Specia, Dhwaj Raj, and Marco Turchi. 2010.
\newblock \href
  {https://link.springer.com/content/pdf/10.1007/s10590-010-9077-2.pdf}
  {Machine translation evaluation versus quality estimation}.
\newblock \emph{Machine Translation}.

\bibitem[{Tiedemann(2012)}]{tiedemann2012parallel}
J{\"o}rg Tiedemann. 2012.
\newblock \href
  {http://www.lrec-conf.org/proceedings/lrec2012/pdf/463_Paper.pdf} {Parallel
  data, tools and interfaces in opus}.
\newblock In \emph{Language Resources and Evaluation Conference (LREC)}.

\bibitem[{Turchi et~al.(2014)Turchi, Anastasopoulos, de~Souza, and
  Negri}]{turchi2014adaptive}
Marco Turchi, Antonios Anastasopoulos, Jos{\'e}~GC de~Souza, and Matteo Negri.
  2014.
\newblock \href {https://www.aclweb.org/anthology/P14-1067.pdf} {Adaptive
  quality estimation for machine translation}.
\newblock In \emph{Annual Meeting of the Association for Computational
  Linguistics (ACL)}.

\bibitem[{Vaswani et~al.(2017)Vaswani, Shazeer, Parmar, Uszkoreit, Jones,
  Gomez, Kaiser, and Polosukhin}]{vaswani2017attention}
Ashish Vaswani, Noam Shazeer, Niki Parmar, Jakob Uszkoreit, Llion Jones,
  Aidan~N Gomez, {\L}ukasz Kaiser, and Illia Polosukhin. 2017.
\newblock \href
  {https://papers.nips.cc/paper/7181-attention-is-all-you-need.pdf} {Attention
  is all you need}.
\newblock In \emph{Conference on Neural Information Processing Systems
  (NeurIPS)}.

\bibitem[{Wang et~al.(2019{\natexlab{a}})Wang, Liu, Wang, Luan, and
  Sun}]{wang2019improving}
Shuo Wang, Yang Liu, Chao Wang, Huanbo Luan, and Maosong Sun.
  2019{\natexlab{a}}.
\newblock \href {https://www.aclweb.org/anthology/D19-1073.pdf} {Improving
  back-translation with uncertainty-based confidence estimation}.
\newblock In \emph{Conference on Empirical Methods in Natural Language
  Processing (EMNLP)}.

\bibitem[{Wang et~al.(2019{\natexlab{b}})Wang, Caswell, and
  Chelba}]{wang-etal-2019-dynamically}
Wei Wang, Isaac Caswell, and Ciprian Chelba. 2019{\natexlab{b}}.
\newblock \href {https://www.aclweb.org/anthology/P19-1123.pdf} {Dynamically
  composing domain-data selection with clean-data selection by
  {``}co-curricular learning{''} for neural machine translation}.
\newblock In \emph{Annual Meeting of the Association for Computational
  Linguistics (ACL)}.

\bibitem[{Wang et~al.(2020)Wang, Tsvetkov, and
  Neubig}]{wang-etal-2020-balancing}
Xinyi Wang, Yulia Tsvetkov, and Graham Neubig. 2020.
\newblock \href {https://www.aclweb.org/anthology/2020.acl-main.754} {Balancing
  training for multilingual neural machine translation}.
\newblock In \emph{Annual Meeting of the Association for Computational
  Linguistics (ACL)}.

\bibitem[{van~der Wees et~al.(2017)van~der Wees, Bisazza, and
  Monz}]{van2017dynamic}
Marlies van~der Wees, Arianna Bisazza, and Christof Monz. 2017.
\newblock \href {https://www.aclweb.org/anthology/D17-1147.pdf} {Dynamic data
  selection for neural machine translation}.
\newblock In \emph{Conference on Empirical Methods in Natural Language
  Processing (EMNLP)}.

\bibitem[{Xia et~al.(2019)Xia, Kong, Anastasopoulos, and
  Neubig}]{xia-etal-2019-generalized}
Mengzhou Xia, Xiang Kong, Antonios Anastasopoulos, and Graham Neubig. 2019.
\newblock \href {https://www.aclweb.org/anthology/P19-1579.pdf} {Generalized
  data augmentation for low-resource translation}.
\newblock In \emph{Annual Meeting of the Association for Computational
  Linguistics (ACL)}.

\bibitem[{Zhang and Zong(2016)}]{zhang2016exploiting}
Jiajun Zhang and Chengqing Zong. 2016.
\newblock \href {https://www.aclweb.org/anthology/D16-1160.pdf} {Exploiting
  source-side monolingual data in neural machine translation}.
\newblock In \emph{Conference on Empirical Methods in Natural Language
  Processing (EMNLP)}.

\bibitem[{Zhang et~al.(2019)Zhang, Shapiro, Kumar, McNamee, Carpuat, and
  Duh}]{zhang2019curriculum}
Xuan Zhang, Pamela Shapiro, Gaurav Kumar, Paul McNamee, Marine Carpuat, and
  Kevin Duh. 2019.
\newblock \href {https://www.aclweb.org/anthology/N19-1189.pdf} {Curriculum
  learning for domain adaptation in neural machine translation}.
\newblock In \emph{Meeting of the North American Chapter of the Association for
  Computational Linguistics (NAACL)}.

\bibitem[{Zheng et~al.(2020)Zheng, Zhou, Huang, Li, Dai, and jun
  Chen}]{Zheng2020MirrorGenerativeNM}
Zaixiang Zheng, Hao Zhou, Shujian Huang, Lei Li, Xin-Yu Dai, and Jia jun Chen.
  2020.
\newblock \href {https://openreview.net/pdf?id=HkxQRTNYPH} {Mirror-generative
  neural machine translation}.
\newblock In \emph{International Conference on Learning Representations
  (ICLR)}.

\bibitem[{Zhou et~al.(2019)Zhou, Zhang, and Hu}]{zhou2019source}
Junpei Zhou, Zhisong Zhang, and Zecong Hu. 2019.
\newblock \href {https://www.aclweb.org/anthology/W19-5411.pdf} {Source:
  Source-conditional elmo-style model for machine translation quality
  estimation}.
\newblock In \emph{Conference on Machine Translation (WMT)}.

\end{thebibliography}

\appendix
\section{Implementation Details}
\begin{itemize}
    \item We use one 11G NVIDIA GTX 1080 GPUs for each experiment. 
    \item The average training time are: about 30 hours for the baseline models and 40 hours for our models.
    \item The number of model parameters is 156.81M.
    \item We use BLEU~\cite{papineni2002bleu} to evaluate the performance of our models,\footnote{\url{https://github.com/moses-smt/mosesdecoder/blob/master/scripts/generic/multi-bleu.perl}} and {\it compare-mt}~\cite{neubig2019compare} to help with the analysis.\footnote{\url{https://github.com/neulab/compare-mt}}
    \item We manually tune the hyperparameters $c_0$ in $[0, 0.1, 0.2]$ and $T$ in $[5, 10, 20]$ in Equation 3, and also the percentage of the selected sentences $p$ in each epoch in $[10\%, 20\%, 30\%, 40\%, 50\%]$. We first set $c_0$ to 0.1, $T$ to $10$ and search for the best $p$, then search for the best $T$, and finally for $c_0$, which takes 11 trials in total.
    \item We follow the instructions on the WMT website to pre-process the data.\footnote{\url{http://data.statmt.org/wmt17/translation-task/preprocessed/de-en/prepare.sh}}
    \item The datasets we use can be downloaded from the WMT website.\footnote{\url{http://www.statmt.org/wmt14}}
\end{itemize}
\end{document}